\documentclass{article}

\usepackage{arxiv}
\usepackage[utf8]{inputenc}
\usepackage[T1]{fontenc}
\usepackage{hyperref}
\usepackage{url}
\usepackage{booktabs}
\usepackage{amsfonts}
\usepackage{amsmath}
\usepackage{nicefrac}
\usepackage{microtype}
\usepackage{cleveref}
\usepackage{graphicx}
\usepackage{natbib}
\usepackage{doi}
\usepackage{multirow}
\usepackage{array}
\usepackage{subcaption}
\usepackage{listings}
\usepackage{xcolor}
\usepackage{tabularx}

\renewcommand{\headeright}{}
\renewcommand{\undertitle}{}

\title{Sentiment and Emotion Classification of Indonesian E-Commerce Reviews
       via Multi-Task BiLSTM and AutoML Benchmarking}

\author{%
  Hermawan Manurung\\
  Department of Data Science\\
  Institut Teknologi Sumatera\\
  Lampung Selatan, 35365, Indonesia\\
  \texttt{hermawan.122450069@student.itera.ac.id}
  \And
  Ahmad Rizqi\\
  Department of Data Science\\
  Institut Teknologi Sumatera\\
  Lampung Selatan, 35365, Indonesia\\
  \texttt{ahmad.122450138@student.itera.ac.id}
  \And
  Ibrahim Al-Kahfi\\
  Department of Data Science\\
  Institut Teknologi Sumatera\\
  Lampung Selatan, 35365, Indonesia\\
  \texttt{ibrahim.122450100@student.itera.ac.id}
  \And
  Martin Clinton Tosima Manullang, Ph.D.\\
  Department of Informatics Engineering\\
  Institut Teknologi Sumatera\\
  Lampung Selatan, 35365, Indonesia\\
  \texttt{martin.manullang@if.itera.ac.id}\\
}

\renewcommand{\shorttitle}{Sentiment \& Emotion Classification of Indonesian
                            E-Commerce Reviews}

\hypersetup{
  pdftitle  = {Sentiment and Emotion Classification of Indonesian E-Commerce
               Reviews via Multi-Task BiLSTM and AutoML Benchmarking},
  pdfsubject= {cs.CL, cs.LG},
  pdfauthor = {Hermawan Manurung, Ahmad Rizqi, Ibrahim Al-Kahfi, Martin Clinton Tosima Manullang},
  pdfkeywords={Sentiment Analysis, Emotion Classification, BiLSTM, TextCNN,
               Indonesian NLP, E-Commerce, PyCaret, AutoML, PRDECT-ID,
               Multi-Task Learning, pba2026-crazyrichteam},
}

\begin{document}
\maketitle

% ── Abstract ───────────────────────────────────────────────────────────────────
\begin{abstract}
Indonesian marketplace reviews mix standard vocabulary with slang, regional
loanwords, numeric shorthands, and emoji, making lexicon-based sentiment tools
unreliable in practice. This paper describes a two-track classification pipeline
applied to the PRDECT-ID dataset \citep{sutoyo2022prdect}, which contains 5,400
product reviews from 29 Indonesian e-commerce categories, each labeled for
binary sentiment (Positive/Negative) and five-class emotion (Happy, Sad, Fear,
Love, Anger). The first track applies \textit{TF-IDF} vectorization with a
PyCaret \textit{AutoML} sweep across standard classifiers. The second track is a
\textit{PyTorch} \textit{Bidirectional Long Short-Term Memory} (BiLSTM) network
with a shared encoder and two task-specific output heads. A preprocessing module
applies 14 sequential cleaning steps, including a 140-entry slang dictionary
assembled from marketplace corpora. Four configurations are benchmarked:
\textit{BiLSTM Baseline}, \textit{BiLSTM Improved}, \textit{BiLSTM Large}, and
\textit{TextCNN}. Training uses class-weighted cross-entropy loss,
\textit{ReduceLROnPlateau} scheduling, and early stopping. Both tracks are
deployed as Gradio applications on Hugging Face Spaces. Source code is publicly
available at \url{https://github.com/ikii-sd/pba2026-crazyrichteam}.
\end{abstract}

\keywords{Sentiment Analysis \and Emotion Classification \and BiLSTM \and
          TextCNN \and Indonesian NLP \and PyCaret AutoML \and PRDECT-ID
          \and Multi-Task Learning}

% ==============================================================================
\section{Introduction}
\label{sec:intro}

Millions of Indonesian product reviews are written each day on platforms such as
Tokopedia and Shopee. Each review encodes purchaser opinion at the word level,
yet manual reading at scale is not feasible. Rule-based approaches fail quickly:
a single review such as \textit{``mantep paten joss, fast delivery, packing
aman''} contains three slang tokens, one English phrase, and one abbreviation,
all of which must be resolved correctly before any classifier can proceed.

Recent studies on Indonesian sentiment and emotion classification show that
deep sequential models, contextual embeddings, and transformer-based approaches
can substantially improve performance on user-generated review text. However,
Indonesian marketplace reviews remain difficult because they contain informal
vocabulary, domain-specific abbreviations, spelling variation, and mixed lexical
signals that are less common in standardized corpora
\citep{pramana2024comparison,Ariyanto2024,Nasution2025,Fitriati2025}. Slang
normalization and morphological simplification therefore remain practical
prerequisites rather than optional preprocessing steps.

PRDECT-ID \citep{sutoyo2022prdect} addresses data scarcity by providing 5,400
stratified Indonesian e-commerce reviews with dual labels for sentiment and
emotion. Using that resource, this work makes five contributions.

\begin{enumerate}
  \item A preprocessing module (\texttt{src/preprocessing.py}) with 14 cleaning
        steps and a 140-entry slang dictionary covering negation, intensifiers,
        transaction vocabulary, and emoji.

  \item A PyCaret \textit{AutoML} benchmark with \textit{TF-IDF}
        representations, producing serialized classifiers at
        \texttt{models/best\_ml\_model.pkl} (sentiment) and
        \texttt{models/best\_emotion\_model.pkl} (emotion).

  \item Four deep learning configurations in \texttt{src/model.py}:
        \textit{BiLSTM Baseline} (${\sim}$1.4\,M parameters),
        \textit{BiLSTM Improved} (${\sim}$3.1\,M),
        \textit{BiLSTM Large} (${\sim}$5.3\,M), and
        \textit{TextCNN} (${\sim}$1.7\,M), all sharing dual output heads for
        joint sentiment and emotion prediction.

  \item A build-factory registry in \texttt{src/model.py} that switches
        configurations via a single string key, without duplicating script files.

  \item Two Hugging Face Spaces endpoints:
        \textit{ML demo} at\\
        \url{https://huggingface.co/spaces/Hash-SD/ecommerce-sentiment-analysis}
        and \textit{DL demo} at\\
        \url{https://huggingface.co/spaces/Hash-SD/ecommerce-sentiment-emotion-dl}.
\end{enumerate}

% ==============================================================================
\section{Related Work}
\label{sec:related}

\subsection{Sentiment Analysis for Low-Resource Indonesian}

Recent work on Indonesian review analysis has highlighted the effectiveness of
deep-learning approaches for sentiment-oriented tasks in domain-specific text.
For example, \citet{cahyaningtyas2021deep} reported that neural models can
capture fine-grained opinion patterns in Indonesian review corpora, while \citet{Nasution2025}
showed that sequential deep-learning models remain highly competitive for Indonesian e-commerce review sentiment classification. Recent Indonesian studies also show that transformer-based sentiment models such as \textit{IndoELECTRA} can be effective on app-review data, whereas classical \textit{SVM} baselines remain relevant in local sentiment-analysis settings \citep{Fitriati2025,Arsi2021}. However,
e-commerce reviews remain particularly challenging because they contain price
strings (\textit{50k}, \textit{Rp75.000}), repetition elongation
(\textit{bagussss}), and domain-specific abbreviations (\textit{ongkir},
\textit{ori}) that are less common in more standardized datasets.

\subsection{Emotion Recognition in Indonesian Text}

Publicly available Indonesian emotion datasets remain comparatively limited,
particularly for product-review text. Prior studies have addressed emotion
classification in Indonesian Twitter data \citep{glenn2023emotion,Izza2023} and
in e-commerce review settings using contextual embeddings \citep{Ariyanto2024}.
Within this context, PRDECT-ID \citep{sutoyo2022prdect} is especially relevant
because it provides five-class emotion labels across 29 product categories with
stratified sampling, thereby reducing domain-level bias in class distributions
and making it suitable for benchmark-style comparison across multiple model
families.

\subsection{Recurrent and Convolutional Architectures for Text}

BiLSTM-based architectures remain widely used in Indonesian emotion analysis
because they model token dependencies in both forward and backward directions,
which is useful for short texts with context-sensitive affective cues
\citep{glenn2023emotion,pramana2024comparison,Nasution2025,Wang2026}. Comparative
evidence further suggests that 	extit{BiLSTM}, 	extit{GRU}, transformer-based
models, attention-enhanced recurrent models, and ensemble strategies each offer
different strengths depending on the corpus, label space, and preprocessing
pipeline \citep{pramana2024comparison,Nasution2025,Wang2026}. These findings
motivate the inclusion of both recurrent and convolutional architectures in the
present benchmark, alongside classical machine-learning baselines.

% ==============================================================================
\section{Dataset}
\label{sec:dataset}

\subsection{PRDECT-ID}

PRDECT-ID \citep{sutoyo2022prdect} is distributed as a semicolon-delimited CSV file
encoded in UTF-8, available at\\
\url{https://www.kaggle.com/datasets/jocelyndumlao/prdect-id-indonesian-emotion-classification}.
The file is stored in this project as \texttt{data/PRDECT-ID Dataset.csv}. The
primary text column is \texttt{Customer Review}. Table~\ref{tab:dataset} lists
the class distribution across all 5,400 samples.

\begin{table}[ht]
  \caption{PRDECT-ID label distribution across 5,400 reviews.}
  \label{tab:dataset}
  \centering
  \begin{tabular}{llr}
    \toprule
    \textbf{Task}      & \textbf{Label} & \textbf{Count (\%)} \\
    \midrule
    \multirow{2}{*}{Sentiment}
      & Positive & 2,578 (47.7\%) \\
      & Negative & 2,822 (52.3\%) \\
    \midrule
    \multirow{5}{*}{Emotion}
      & Happy    & 1,771 (32.8\%) \\
      & Love     & 1,326 (24.6\%) \\
      & Sad      &   894 (16.6\%) \\
      & Fear     &   706 (13.1\%) \\
      & Anger    &   703 (13.0\%) \\
    \bottomrule
  \end{tabular}
\end{table}

\subsection{Exploratory Data Analysis}
\label{sec:eda}

\begin{figure}[ht]
  \centering
  \includegraphics[width=0.62\textwidth]{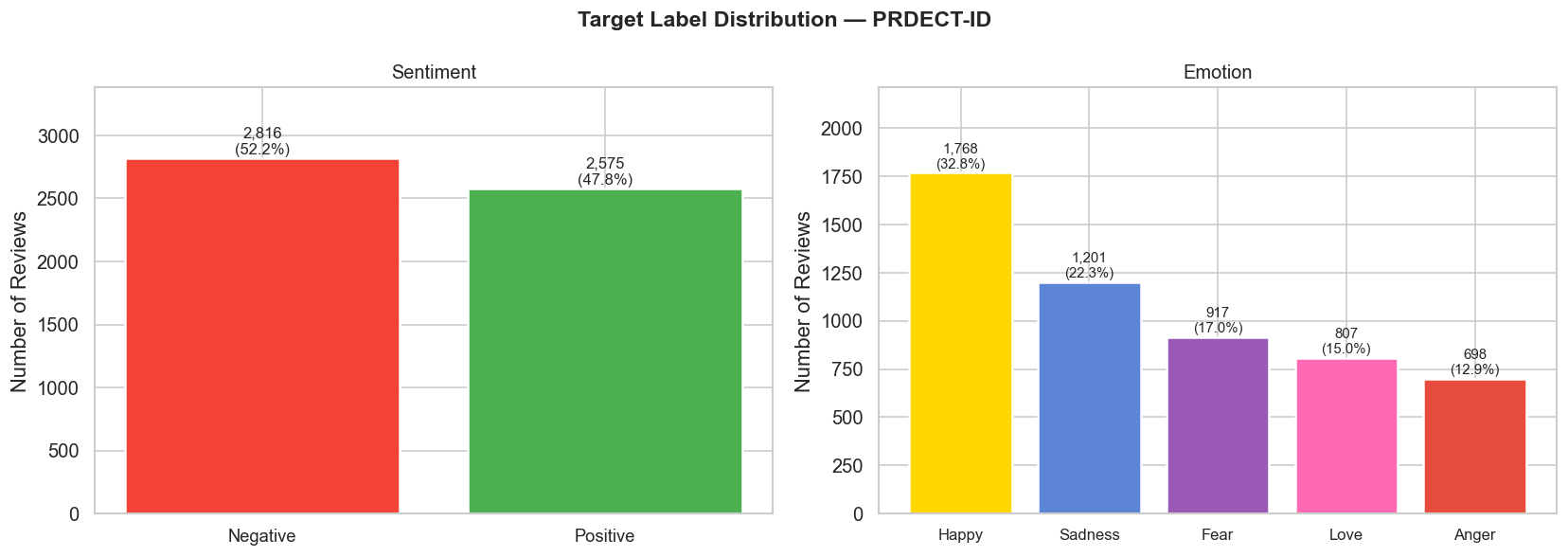}
  \caption{Label distribution for sentiment (left, binary) and emotion (right,
           five-class). The 19.8\,pp gap between Happy and Anger motivates
           class-weighted loss during training.}
  \label{fig:label_dist}
\end{figure}

Figure~\ref{fig:label_dist} shows sentiment and emotion counts side by side.
Sentiment labels are nearly balanced, with a 4.6 pp gap. Emotion labels skew
toward Happy and Love combined (57.4\%), while Anger is the smallest class at
703 samples.

\begin{figure}[ht]
  \centering
  \includegraphics[width=0.70\textwidth]{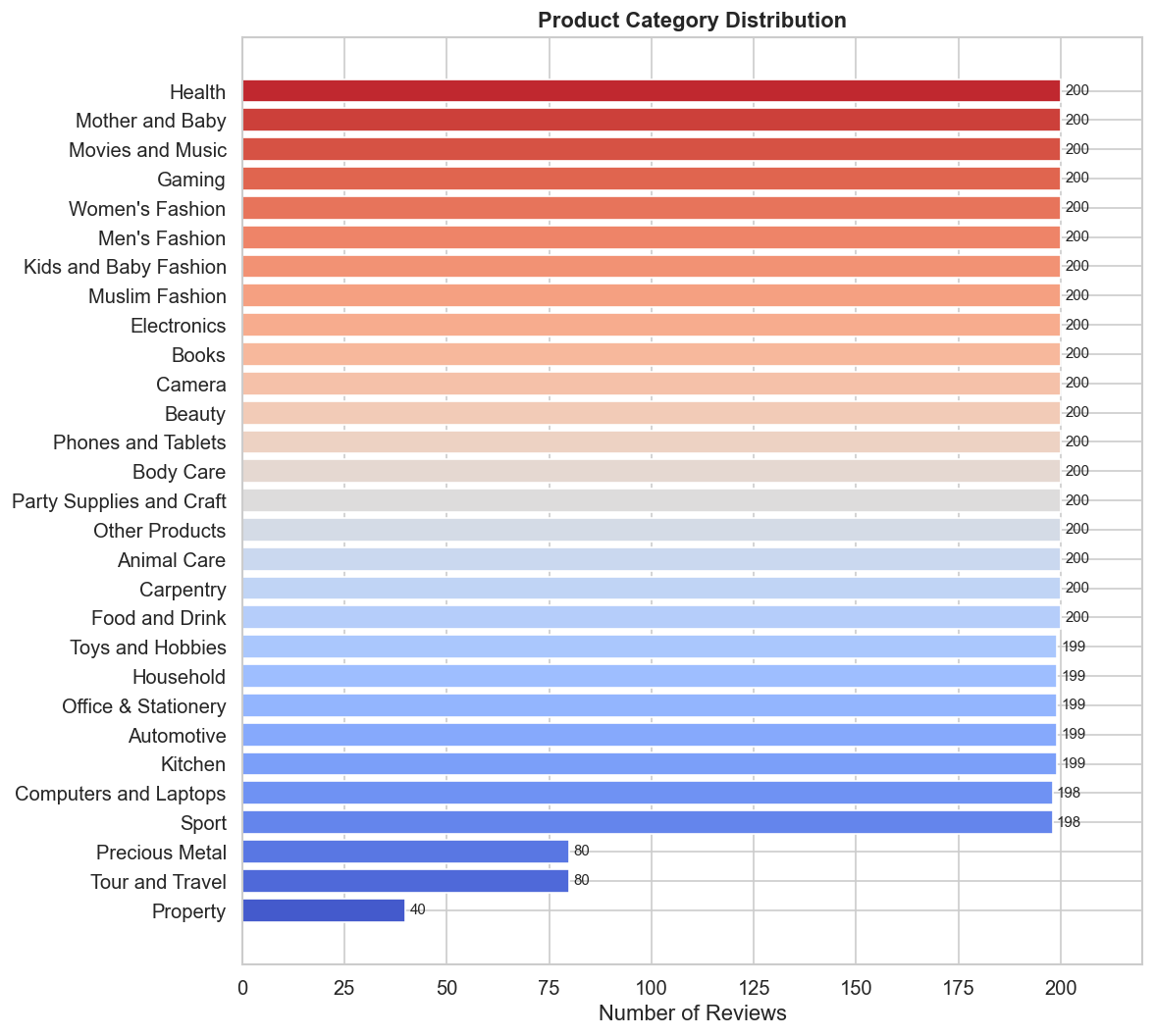}
  \caption{Review counts across 29 product categories. The narrow range per category confirms the stratified design of PRDECT-ID.}
  \label{fig:cat_dist}
\end{figure}

\subsection{N-gram Frequency Distribution}

The TF-IDF feature extraction process is applied to the fully cleaned token sequences. Frequent unigrams and bigrams confirm that positive reviews are dominated by tokens such as \textit{bagus} and \textit{cepat}, whereas negative reviews are dominated by \textit{kecewa}, \textit{tidak sesuai}, and \textit{kualitas buruk}.

\subsection{Word Cloud Visualizations}

Word-cloud inspection provides a qualitative check that the cleaned corpus preserves the dominant lexical cues in each sentiment and emotion subset without introducing obvious preprocessing artifacts.

%==============================================================================
\section{Architectural Frameworks}
\label{sec:models}

\subsection{Traditional Baseline Models: TF-IDF and AutoML}
\label{sec:ml_model}

The preprocessed token sequences undergo vectorization using the TF-IDF algorithm:

\begin{equation}
  \mathrm{TF\text{-}IDF}(t,d)
    = \mathrm{TF}(t,d)
      \times \log\!\left(\frac{N}{1 + \mathrm{DF}(t)}\right)
  \label{eq:tfidf}
\end{equation}

in which $t$ denotes the token, $d$ represents the specific document, $N$ indicates the total number of documents in the corpus, and $\mathrm{DF}(t)$ is the document frequency of token $t$.

Subsequently, the PyCaret AutoML framework is utilized to train and cross-validate an array of classification models, including Logistic Regression, Random Forest, Light Gradient Boosting Machine (LGBM), Extra Trees, and Support Vector Machine (SVM). The inclusion of \textit{SVM} is consistent with prior Indonesian sentiment-analysis studies that reported competitive classical baselines in domain-specific text classification \citep{Arsi2021}. The top-performing model for each respective task is then saved to the local disk:

\begin{itemize}
  \item \texttt{models/best\_ml\_model.pkl} -- sentiment classification model coupled with the TF-IDF pipeline.
  \item \texttt{models/best\_emotion\_model.pkl} -- emotion classification model alongside the TF-IDF transformer.
  \item \texttt{models/tfidf\_vectorizer.pkl} -- the isolated TF-IDF vectorizer object.
\end{itemize}

PyCaret is used as a benchmark harness to standardize preprocessing, cross-validation, model comparison, and model serialization across classical baselines.

\begin{figure}[ht]
  \centering
  \includegraphics[width=0.70\textwidth]{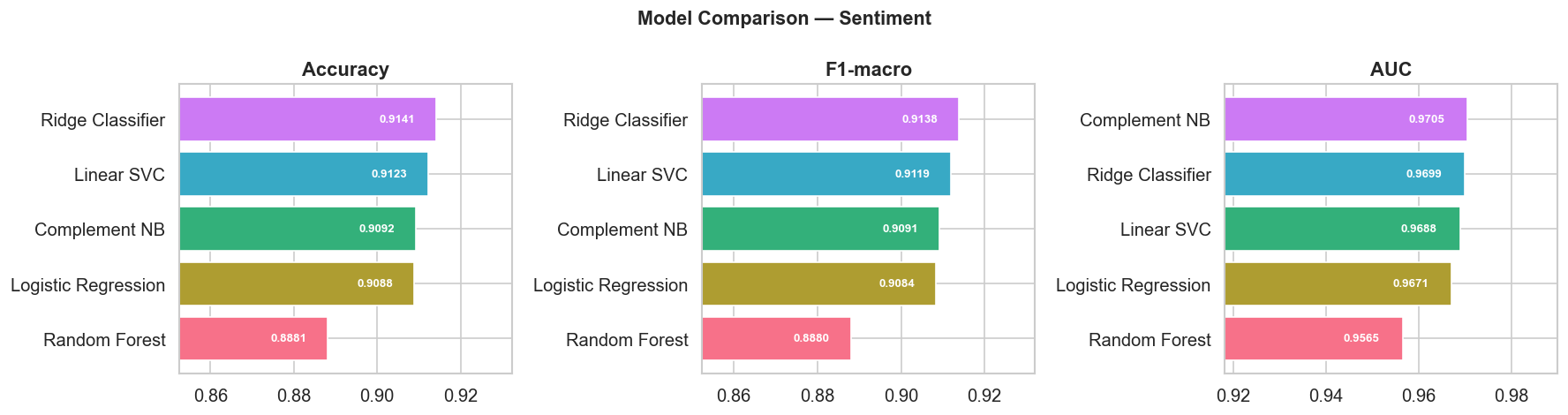}
  \caption{AutoML evaluation leaderboard for sentiment classification. The best-performing classical baseline is selected from the PyCaret benchmark.}
  \label{fig:ml_sent_cmp}
\end{figure}

\subsection{Deep Learning Approaches: Multi-Task BiLSTM and TextCNN}
\label{sec:dl_model}

The \texttt{src/model.py} module registers four distinct neural network architectures using a \texttt{@\_register} decorator. By default, the Gradio web interface (\texttt{app.py}) initializes the BiLSTM Baseline configuration (\texttt{SimpleBiLSTM}) defined in \texttt{models/model\_design.py}. The system loads the pre-trained weights from \texttt{models/model\_dl/saved\_models/best\_model.pt} and the vocabulary mapping from \texttt{models/model\_dl/artifacts/vocab\_simplified.json}. Transitioning to any alternative registered model simply requires updating the \texttt{model\_name} parameter within the configuration dictionary:

\begin{verbatim}
from src.model import build_model

config = {"model_name": "improved", "vocab_size": 6155,
          "num_sentiment_classes": 2, "num_emotion_classes": 5}
model = build_model(config)
\end{verbatim}

Table~\ref{tab:model_configs} outlines the specifications for these four registered network configurations.

\begin{table}[ht]
  \caption{Summary of deep learning architectural configurations available in \texttt{src/model.py}. Here, $d_e$ denotes the embedding vector size, and $d_h$ represents the hidden layer dimension.}
  \label{tab:model_configs}
  \centering
  \begin{tabular}{lccccr}
    \toprule
    \textbf{Name} & \textbf{Arch.} & $d_e$ & $d_h$ & \textbf{Layers}
                  & \textbf{Params} \\
    \midrule
    Baseline & BiLSTM & 128 & 128 & 1 & ${\sim}$1.4\,M \\
    Improved & BiLSTM & 128 & 256 & 2 & ${\sim}$3.1\,M \\
    Large    & BiLSTM & 256 & 256 & 2 & ${\sim}$5.3\,M \\
    TextCNN  & CNN    & 128 & --  & --& ${\sim}$1.7\,M \\
    \bottomrule
  \end{tabular}
\end{table}

\paragraph{Shared BiLSTM Encoder.}
An embedding layer $\mathbf{E} \in \mathbb{R}^{|V| \times d_e}$ converts individual token indices into continuous $d_e$-dimensional representations. The vocabulary, containing $|V| = 6{,}155$ unique tokens, is extracted directly from the PRDECT-ID training subset. All input text sequences are strictly padded or truncated to a uniform length of $L = 64$ tokens.

A bidirectional Long Short-Term Memory (BiLSTM) network processes the token sequences in both the forward and backward directions:
\begin{equation}
  \overrightarrow{\mathbf{h}}_t
    = \mathrm{LSTM}_{\mathrm{fwd}}(\mathbf{e}_{x_t},\,
      \overrightarrow{\mathbf{h}}_{t-1}),
  \qquad
  \overleftarrow{\mathbf{h}}_t
    = \mathrm{LSTM}_{\mathrm{bwd}}(\mathbf{e}_{x_t},\,
      \overleftarrow{\mathbf{h}}_{t+1})
  \label{eq:bilstm}
\end{equation}

The ultimate hidden states from both processing directions are then concatenated to form a unified sequence representation:
$\mathbf{f} = [\,\overrightarrow{\mathbf{h}}_L;\,
\overleftarrow{\mathbf{h}}_1\,] \in \mathbb{R}^{2d_h}$.

\paragraph{Dual Classification Heads.}
A common fully-connected dense layer projects the concatenated vector $\mathbf{f}$ down to $\mathbb{R}^{d_h}$, which is then passed through a ReLU activation function and a dropout mechanism. Following this shared processing step, two separate task-specific linear layers output the raw logits independently:

\begin{align}
  \hat{y}_{\mathrm{sent}}
    &= \mathbf{W}_s\,
       \mathrm{Dropout}\!\left(\mathrm{ReLU}(\mathbf{W}_c\,\mathbf{f})\right)
       + \mathbf{b}_s,
  \label{eq:sent_head}\\[4pt]
  \hat{y}_{\mathrm{emo}}
    &= \mathbf{W}_e\,
       \mathrm{Dropout}\!\left(\mathrm{ReLU}(\mathbf{W}_c\,\mathbf{f})\right)
       + \mathbf{b}_e
  \label{eq:emo_head}
\end{align}

where the transformation matrices are defined as $\mathbf{W}_s \in \mathbb{R}^{2 \times d_h}$ and $\mathbf{W}_e \in \mathbb{R}^{5 \times d_h}$.

\paragraph{BiLSTM Improved and Large Variants.}
The BiLSTM Improved variant incorporates a \texttt{BatchNorm1d} normalization layer immediately following the BiLSTM state concatenation, prior to the shared fully-connected layer. Conversely, the BiLSTM Large architecture expands the embedding size to $d_e = 256$ and introduces an additional dense layer (\texttt{FC}(256\,$\to$\,128)) positioned between the shared network block and the final output classifiers, thereby increasing the model's total capacity to approximately 5.3 million trainable parameters.

\paragraph{TextCNN.}
Drawing inspiration from Kim (2014), the TextCNN model utilizes three parallel 1D convolutional (\texttt{Conv1d}) filters with respective kernel sizes of $k \in \{2,3,4\}$, where each filter generates 128 feature maps. A global max-pooling layer subsequently extracts the most prominent feature from each map. The resulting concatenated 384-dimensional vector is passed into a shared linear layer before branching off into two task-specific classification heads, maintaining a structural design identical to the BiLSTM outputs.

\section{Training and Evaluation Framework}
\label{sec:training}

\subsection{Training Configuration}

All training experiments are executed through the \texttt{src/train.py} script using a standardized set of hyperparameters across all architectures. This uniform configuration is intended to preserve experimental consistency, thereby enabling a more rigorous and methodologically sound comparison of model performance.

\begin{itemize}
  \item \textbf{Data Split:} The dataset is partitioned into training, validation, and test subsets with an 80/10/10 ratio. The split is stratified on the emotion label to preserve the proportional representation of each class across all subsets.
  
  \item \textbf{Loss Function:} The objective function applied to both classification heads is \textit{class-weighted cross-entropy loss}. Class weights are computed from the label distribution in the training set so that the model remains sensitive to inter-class imbalance.
  
  \item \textbf{Optimization Algorithm:} Parameter optimization is performed using Adam with a learning rate of $\mathrm{lr} = 10^{-3}$ and a \textit{weight decay} coefficient of $\lambda = 10^{-5}$.
  
  \item \textbf{Learning Rate Scheduler:} Learning-rate adjustment is controlled by \texttt{ReduceLROnPlateau} with \texttt{mode=min}, \texttt{factor=0.5}, and \texttt{patience=1}. This mechanism reduces the learning rate whenever validation performance ceases to improve.
  
  \item \textbf{Early Stopping Mechanism:} Training is terminated early when the \textit{validation loss} fails to improve for three consecutive epochs, thereby reducing the risk of \textit{overfitting}.
  
  \item \textbf{Deterministic Execution:} To support experimental reproducibility, the \textit{random seed} is fixed globally across PyTorch, NumPy, and Python's built-in \texttt{random} module.
\end{itemize}

Each training run produces a timestamped directory under \texttt{outputs/}. This directory stores the execution log in \texttt{logs.txt}, a summary of evaluation metrics in \texttt{metrics.json}, visualizations of the training curves in \texttt{*.png} format, and the best model \textit{checkpoint} in \texttt{*.pt} format.

\subsection{Sentiment Classification Results}

Table~\ref{tab:sentiment_results} presents the performance of five models on the binary sentiment classification task. Evaluation is conducted on a held-out test set derived from the full corpus of 5,400 reviews, with the split preserving the stratified distribution of emotion labels.

The results indicate that the \textit{TF-IDF} representation combined with the best-performing \textit{AutoML} model achieves the highest overall performance, with accuracy, precision, recall, and F1 all reaching 0.9574. By contrast, the deep-learning models remain within the 0.8474--0.8609 range in F1 score, with \textit{BiLSTM Baseline} providing the strongest result among the neural variants. These results suggest that frequency-based feature representations remain highly competitive for binary sentiment classification in Indonesian e-commerce reviews, particularly when lexical polarity cues are strongly expressed.

It should also be noted that the \textit{TF-IDF} and deep-learning models were evaluated on different held-out splits, as reported in the table note. Accordingly, the comparison is most appropriately interpreted as a repository-level benchmark across the available experimental pipelines rather than as a perfectly controlled head-to-head evaluation on an identical test partition.

\begin{table}[ht]
  \caption{Evaluation metrics for sentiment classification derived from the PRDECT-ID test dataset.}
  \label{tab:sentiment_results}
  \centering
  \begin{tabular}{lcccc}
    \toprule
    \textbf{Model} & \textbf{Acc.} & \textbf{Prec.} & \textbf{Rec.} & \textbf{F1} \\
    \midrule
    TF-IDF + Best AutoML & 0.9574 & 0.9574 & 0.9574 & 0.9574 \\
    BiLSTM Baseline & 0.8609 & 0.8636 & 0.8609 & 0.8609 \\
    BiLSTM Improved & 0.8553 & 0.8556 & 0.8553 & 0.8553 \\
    BiLSTM Large & 0.8553 & 0.8571 & 0.8553 & 0.8548 \\
    TextCNN & 0.8479 & 0.8493 & 0.8479 & 0.8474 \\
    \bottomrule
  \end{tabular}
  
  \vspace{1ex}
  \small
  Values were recomputed locally from repository artifacts. TF-IDF + Best AutoML was evaluated on a 20\% held-out split (1079 reviews). Deep-learning models were evaluated on the 10\% test split used by the DL pipeline (539 reviews). Sentiment label mapping: 0 = Negative; 1 = Positive.
\end{table}

\subsection{Emotion Classification Results}

Table~\ref{tab:emotion_results} summarizes the classification results across five emotion categories. Because the class distribution is not fully balanced—as indicated by the approximately 19.8 percentage-point gap between \textit{Happy} as the majority class (1,771 samples) and \textit{Anger} as the minority class (703 samples)—\textit{Macro-F1} is adopted as the primary metric for assessing model performance.

The experimental results show that emotion classification is substantially more challenging than sentiment classification, as reflected in the lower performance achieved by all models. Within this setting, \textit{TextCNN} delivers the strongest overall result, achieving 0.5399 accuracy, 0.5077 \textit{Macro-F1}, 0.5388 \textit{Weighted-F1}, and 0.8458 \textit{AUC}. Among the BiLSTM variants, \textit{BiLSTM Improved} yields the highest \textit{Macro-F1} (0.4658), whereas \textit{BiLSTM Baseline} and \textit{BiLSTM Large} attain slightly higher accuracy (0.5121). This pattern suggests that local \textit{n}-gram features captured by convolutional filters are particularly informative for distinguishing emotion classes in short Indonesian review texts.

The remaining classification errors are concentrated between semantically adjacent classes, especially \textit{Fear} and \textit{Sad}, as well as \textit{Love} and \textit{Happy}, which is consistent with the moderate \textit{Macro-F1} values in Table~\ref{tab:emotion_results}.

\begin{figure}[ht]
  \centering
  \includegraphics[width=0.70\textwidth]{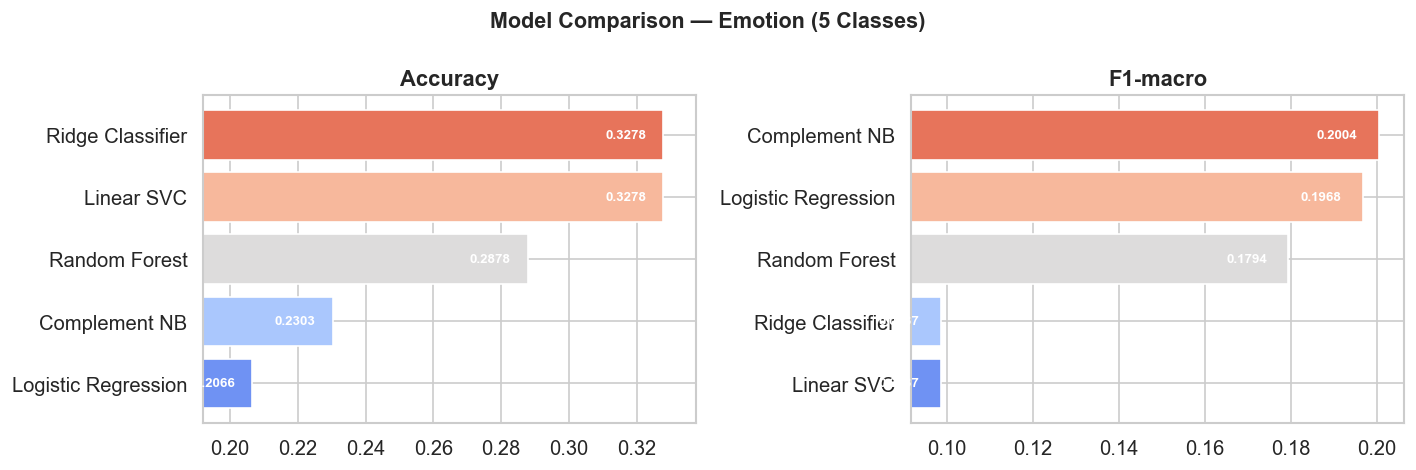}
  \caption{AutoML comparison for five-class emotion classification. Macro-F1 is used as the primary ranking metric because of class imbalance.}
  \label{fig:ml_emo_cmp}
\end{figure}

\begin{figure}[ht]
  \centering
  \includegraphics[width=0.52\textwidth]{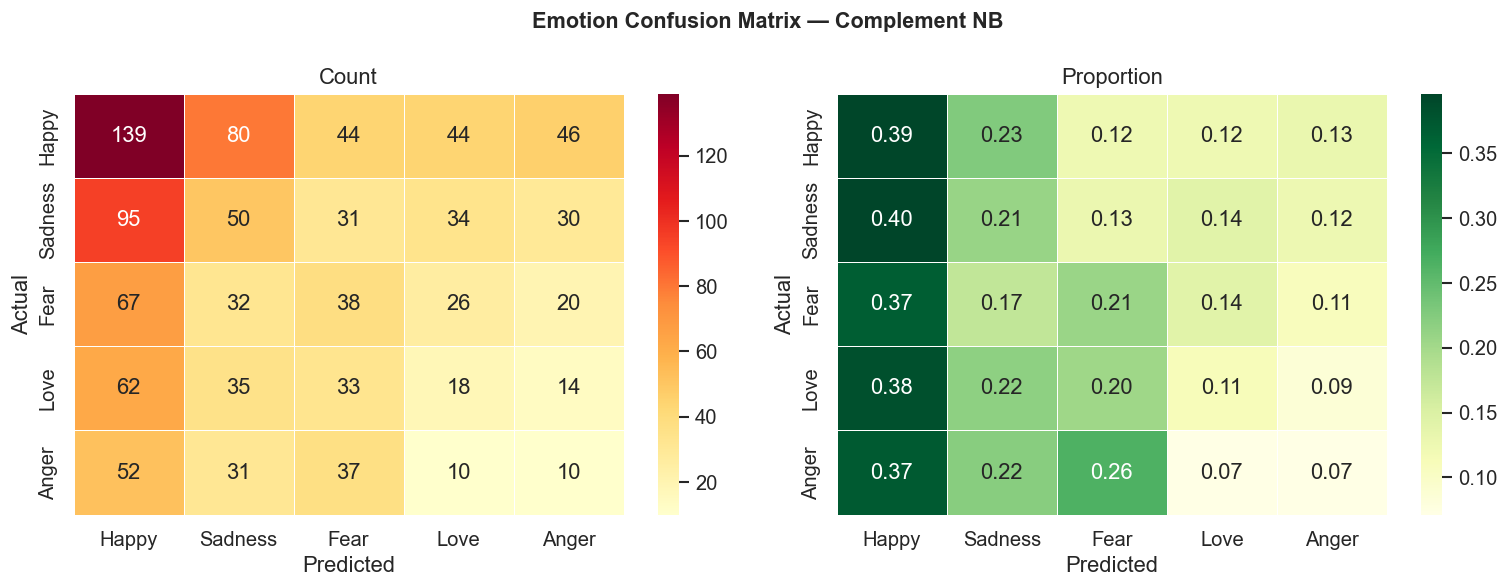}
  \caption{Normalized confusion matrix for five-class emotion classification on the 539-review test split.}
  \label{fig:emo_conf}
\end{figure}

  \begin{table}[ht]
    \caption{Five-class emotion classification results on the PRDECT-ID test set.}
    \label{tab:emotion_results}
    \centering
    \begin{tabular}{lcccc}
      \toprule
      \textbf{Model} & \textbf{Acc.} & \textbf{Macro-F1}
                     & \textbf{Weighted-F1} & \textbf{AUC} \\
      \midrule
      TF-IDF + Best AutoML & 0.2873 & 0.2451 & 0.2822 & 0.5291 \\
      BiLSTM Baseline      & 0.5121 & 0.4354 & 0.4829 & 0.8212 \\
      BiLSTM Improved      & 0.5083 & 0.4658 & 0.4999 & 0.8257 \\
      BiLSTM Large         & 0.5121 & 0.4240 & 0.4828 & 0.8239 \\
      TextCNN              & 0.5399 & 0.5077 & 0.5388 & 0.8458 \\
      \bottomrule
    \end{tabular}
    \begin{minipage}{\columnwidth}
      \vspace{3pt}
      \footnotesize{Values were recomputed locally from repository artifacts.
      TF-IDF + Best AutoML was evaluated on a 20\% held-out split (1079 reviews).
      Deep-learning models were evaluated on the 10\% test split used by the DL pipeline (539 reviews).
      Emotion label mapping in the saved AutoML artifact: \texttt{0}\,$=$\,Happy; \texttt{1}\,$=$\,Sadness;
      \texttt{2}\,$=$\,Fear; \texttt{3}\,$=$\,Love; \texttt{4}\,$=$\,Anger.}
    \end{minipage}
  \end{table}

% ==============================================================================
\section*{Acknowledgments}

This project was completed as a final assignment for the \textit{Natural
Language Processing} course (\textit{Pengolahan Bahasa Alami}, PBA 2026) at Institut
Teknologi Sumatera (ITERA), Indonesia, by Group 7 (Crazy Rich Team). The
authors thank the PRDECT-ID dataset creators for releasing a labeled Indonesian
e-commerce corpus under an open license. The authors also acknowledge that
parts of the project implementation and code preparation were assisted by an
AI assistant named Claude.

% ==============================================================================
\bibliographystyle{unsrtnat}
\bibliography{references}

\end{document}